\documentclass[letterpaper]{article} 
\usepackage{aaai23}  
\usepackage{times}  
\usepackage{helvet}  
\usepackage{courier}  
\usepackage[hyphens]{url}  
\usepackage{graphicx} 
\usepackage{natbib}  
\usepackage{caption} 
\usepackage{algorithm}
\usepackage{algorithmic}

\def\eg{\emph{e.g.}, }
\def\etal{{\em et al.~}}

\usepackage{newfloat}
\usepackage{listings}
\DeclareCaptionStyle{ruled}{labelfont=normalfont,labelsep=colon,strut=off} 
\floatstyle{ruled}
\newfloat{listing}{tb}{lst}{}
\floatname{listing}{Listing}
\usepackage{multirow}
\usepackage{threeparttable}

\title{FoPro: Few-Shot Guided Robust Webly-Supervised Prototypical Learning}
\author{
    Yulei Qin,\equalcontrib\textsuperscript{\rm 1}
    Xingyu Chen,\equalcontrib\textsuperscript{\rm 1}
    Chao Chen,\textsuperscript{\rm 1}
    Yunhang Shen,\textsuperscript{\rm 1}
    Bo Ren,\textsuperscript{\rm 1}\\
    Yun Gu,\textsuperscript{\rm 2}
    Jie Yang,\textsuperscript{\rm 2}
    Chunhua Shen\textsuperscript{\rm 3}
}
\affiliations{
    \textsuperscript{\rm 1}Tencent YouTu Lab,
    \textsuperscript{\rm 2}Shanghai Jiao Tong University,
    \textsuperscript{\rm 3}Zhejiang University\\
    \{yuleiqin, harleychen, aaronccchen, odysseyshen, timren\}@tencent.com,\\
    yungu@ieee.org,
    jieyang@sjtu.edu.cn,
    chunhuashen@zju.edu.cn
}

\usepackage{bibentry}

\begin{document}

\maketitle

\begin{abstract}
Recently, webly supervised learning (WSL) has been studied to leverage numerous and accessible data from the Internet.
Most existing methods focus on learning noise-robust models from web images while neglecting the performance drop caused by the differences between web domain and real-world domain.
However, only by tackling the performance gap above can we fully exploit the practical value of web datasets.
To this end, we propose a Few-shot guided Prototypical (FoPro) representation learning method, which only needs a few labeled examples from reality and can significantly improve the performance in the real-world domain.
Specifically, we initialize each class center with few-shot real-world data as the ``realistic" prototype.
Then, the intra-class distance between web instances and ``realistic" prototypes is narrowed by contrastive learning.
Finally, we measure image-prototype distance with a learnable metric.
Prototypes are polished by adjacent high-quality web images and involved in removing distant out-of-distribution samples.
In experiments, FoPro is trained on web datasets with a few real-world examples guided and evaluated on real-world datasets.
Our method achieves the state-of-the-art performance on three fine-grained datasets and two large-scale datasets.
Compared with existing WSL methods under the same few-shot settings, FoPro still excels in real-world generalization.
Code is available at \url{https://github.com/yuleiqin/fopro}.
\end{abstract}

\section{Introduction}

The past decade has witnessed a revolution in computer vision with the advent of large-scale labeled datasets (\eg{ImageNet \cite{deng2009imagenet}}).
However, a large collection of data are sometimes inaccessible,
let alone the time-consuming and expensive annotations.
On the contrary, there are abundant weakly labeled images on the Internet.
Therefore, webly supervised learning (WSL) has attracted growing attention from researchers \cite{krause2016unreasonable, kaur2017combining, kolesnikov2019large, zhang2020web, tu2020learning, liu2021exploiting, zhang2021understanding}.

\begin{figure}[htbp]
\centering
\includegraphics[width=0.9\columnwidth]{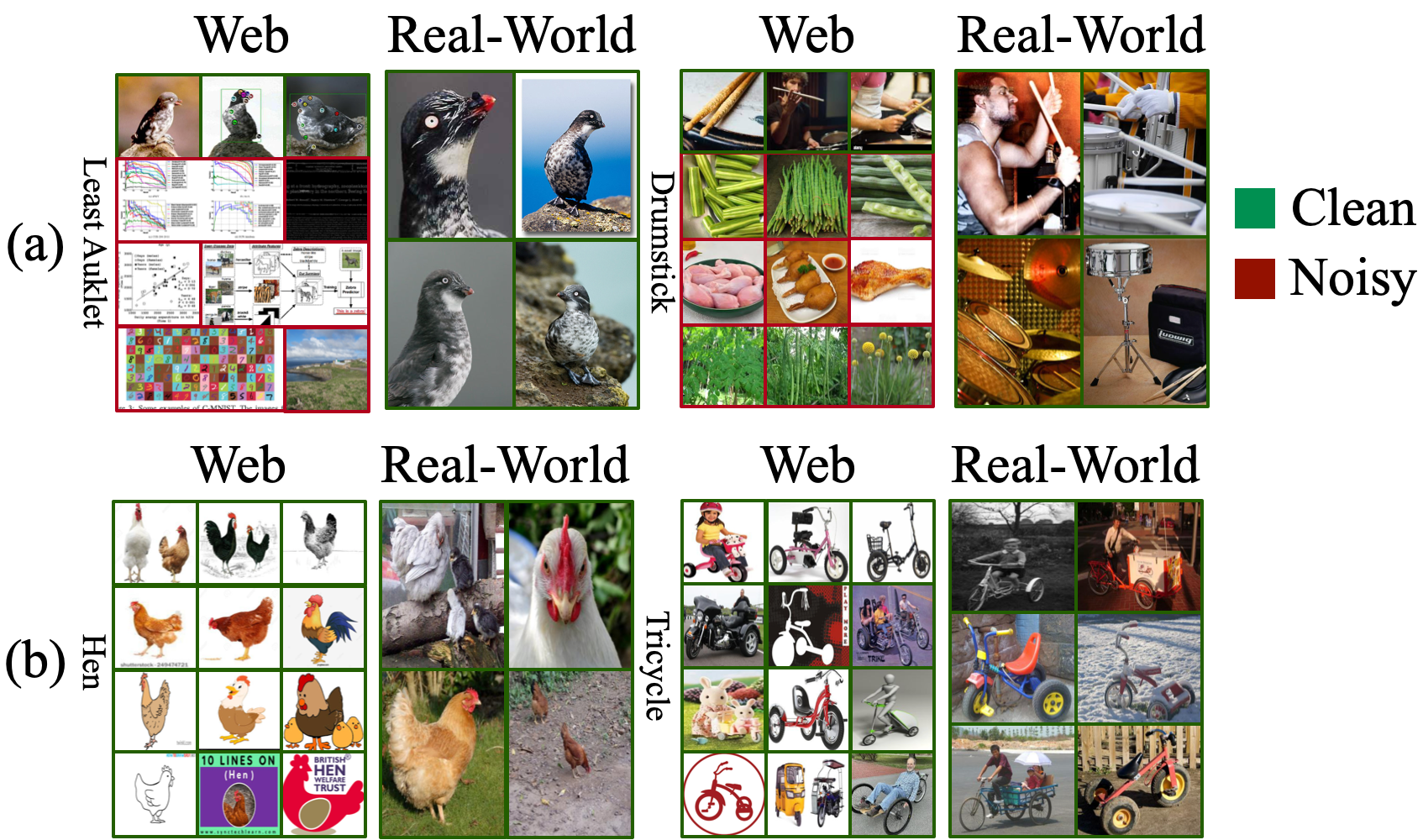}
\caption{Differences between web and real-world images. (a) Web dataset noise. (b) Web dataset bias.
}
\label{fig:challenges}
\end{figure}

Queries and tags are directly used as weak labels without verification,
bringing about a considerable proportion of noises in web datasets (\eg{20\% in JMT-300M \cite{sun2017revisiting}, 34\% in WebVision \cite{li2017webvision}, and 32\% in WebFG496 \cite{sun2021webly}}).
As shown in Fig. \ref{fig:challenges}(a), various noises include label flipping errors,
semantic ambiguity of polysemy queries,
and outliers of unknown categories.
To alleviate their effect, prior knowledge such as neighbor density \cite{guo2018curriculumnet},
reference clean sets \cite{jiang2018mentornet, lee2018cleannet},
and side information \cite{zhou2020large, cheng2020weakly} is explored for label correction and sample selection.
Recently, Li \etal{\cite{li2020mopro}} develop a self-supervised method with representative class prototypes (MoPro) to achieve satisfying performance.

Most existing WSL methods are merely concerned with noise reduction.
They ignore the model degradation in real-world scenarios because the performance on web domain testing sets is emphasized in previous model assessments.
Domain gaps exist between images crawled from the web (\eg{advertising photos, artworks, and rendering}) and those captured in reality (see Fig. \ref{fig:challenges}(b)).
In this case, the better fitting of web images counteractively leads to worse generalization on practical applications.
Few studies try to tackle such performance gaps by domain adaptation methods.
For example, Xu \etal{\cite{xu2016webly}} distill knowledge from the web domain to the real-world domain.
Niu \etal{\cite{niu2015visual}} fine-tune pretrained models on real-world datasets.
However, both of them need plenty of labeled data in the target domain, which impedes practicability.

\begin{figure*}[ht]
\centering
\includegraphics[width=0.9\textwidth]{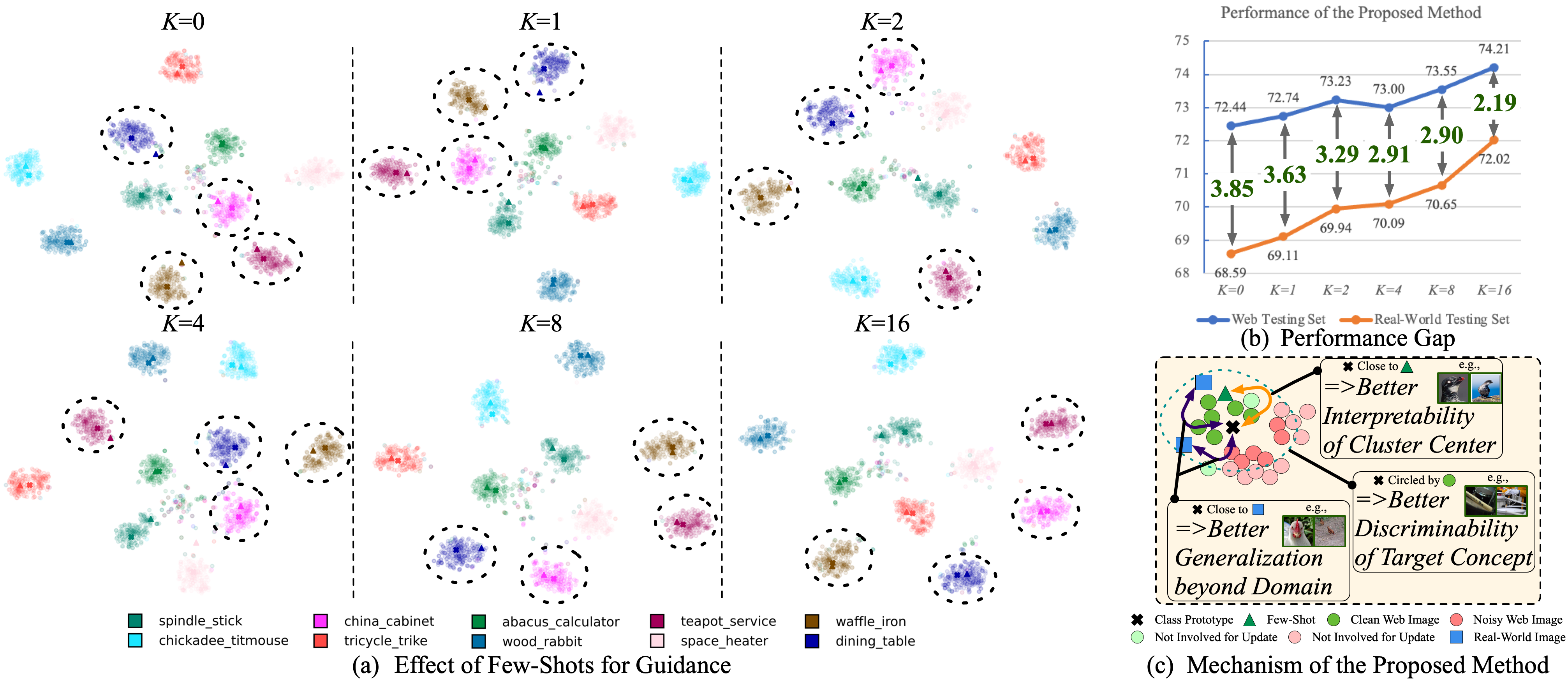}
\caption{(a) The t-SNE \cite{van2008visualizing} of the low-dimensional embeddings of web images substantiates that with the increase of $K$, class prototypes ($\times$) are regularized to approach few shots ($\Delta$) with dense intra-class and isolated inter-class distribution. (b) The diminished performance gap between the testing results of web (WebVision1k) and real-world (ImageNet1k) images confirms that FoPro takes full advantage of few shots to improve its generalization beyond the web domain, making web data truly useful in learning representations for actuality. (c) FoPro estimates noise-robust prototypes to pull instances nearby closer. Noisy samples are filtered by assessing their relation with prototypes. Only clean ones update prototypes in return. FoPro achieves better interpretability, discriminability, and generalization. Best viewed magnified.}
\label{fig:proposed}
\end{figure*}

Unlike the methods above, our objective is to cost-efficiently mine web data for real-world applications.
We handle both the noise and domain gap by resorting to a few human-labeled samples for guidance on \textit{whom to learn from} and \textit{what to learn}.
In our setting, clean labeled examples are too scarce to train or fine-tune a deep model, and therefore alternative methods need to be developed in response.

To this end, we propose a robust prototypical representation learning method with noisy web data and a few clean examples (see Fig.\ref{fig:proposed}).
Motivated by the anchoring role of class prototypes \cite{li2020prototypical, li2020mopro}, we introduce \textbf{F}ew-sh\textbf{o}t guided \textbf{Pro}totypes, termed as \textbf{FoPro}, to effectively deal with noise and domain gap.
Technically, we project features of the penultimate layer of a classification model to low-dimensional embeddings.
The critical problem is how to \textit{formulate a class-representative and domain-generalized prototype} in the embedding space without being deviated by the dominating noises.
Due to noise memorization \cite{arpit2017closer}, simply averaging over instances with high prediction confidence does not promise a noise-robust estimation.
Consequently, we first initialize each class prototype with realistic few shots as the cluster center.
Secondly, intra-class distance is shortened by contrastive learning between web instances and prototypes.
Then, high-quality web examples are involved in polishing prototypes to improve discriminability.
Simultaneously, high similarity between prototypes and few shots is regularized to maximize the interpretability and generalizability of prototypes.
Finally, we quantify the compatibility between instances and prototypes by the proposed relation module for sample selection and correction,
which benefits prototype update in the next iteration.
Specifically, the relation module learns a flexible and transferable metric to assess if a web image corresponds to its label.
Besides, we set siamese encoders \cite{he2020momentum} and prototypes are only updated by the momentum encoder in a smooth and progressive way.

Our contributions can be summarized as follows:
\begin{itemize}
    \item We propose a new few-shot learning setting in WSL with abundant noisy web images and a few real-world images, which aims to improve the performance of WSL for real-world applications in a cost-efficient way.
    \item We present a new method under the setting above called FoPro, which simultaneously solves noise and data bias in an end-to-end manner.
    Experimental results show that our method can significantly improve the performance in real-world benchmark datasets.
    \item We propose a new relation module for label noise correction.
    It outperforms existing methods that use fixed metrics (\eg{cosine distance}) by evaluating instance-prototype similarity with a learnable metric.
    \item Extensive experiments on the fine-grained WebFG496 and the large-scale WebVision1k datasets confirm the superiority of FoPro over the state-of-the-art (SOTA) methods. Performance under the increasing $K$-shot settings demonstrates that FoPro utilizes few shots wisely to bridge the gap towards real-world applications.
\end{itemize}

\section{Related Work}

\subsection{Webly Supervised Learning}

WSL aims to leverage vast but weakly-annotated web resources.
Previous works utilize web images for tasks including classification \cite{bergamo2010exploiting, wu2021ngc, yao2017exploiting, yao2020bridging}, detection \cite{divvala2014learning, shen2020noise}, and segmentation \cite{shen2018bootstrapping, jin2017webly}.

Recently, noise cleaning methods such as self-contained confidence (SCC) \cite{yang2020webly} and momentum prototype (MoPro) \cite{li2020mopro} are proposed to improve representation learning in WSL.
SCC balances two supervision sources from web labels and predicted labels by introducing instance-wise confidence.
MoPro targets model pretraining for several down-streaming tasks by combining self-supervised and webly-supervised techniques.
Specifically, MoPro is closely related to ours since the contrast between instances and prototypes is used to learn discriminative features.
Different from MoPro, we formulate a brand-new setting where a few samples labeled by experts are available.
To assure that class prototypes are not misled by noise, an implicit constraint on distribution is achieved by enforcing high similarity between prototypes and few shots.
Furthermore, we estimate the relation score between instances and prototypes to correct labels and discard out-of-distribution (OOD) samples.

\subsection{Learning from Noisy Labels}

Labels in human-annotated datasets can still be noisy due to lack of expert domain knowledge \cite{song2022learning}.
To prevent deep models from overfitting noisy labels, several studies have been conducted and can be categorized as:
1) robust architecture (\eg{noise transition layer \cite{chen2015webly} and probability model \cite{xiao2015learning}});
2) regularization techniques (\eg{label smoothing \cite{pereyra2017regularizing} and mix-up \cite{zhang2018mixup}});
3) robust losses (\eg{MAE \cite{ghosh2017robust} and GCE \cite{zhang2018generalized}});
4) loss refinement (\eg{reweighting \cite{wang2017multiclass} and bootstrapping \cite{reed2015training}});
5) sample selection (\eg{multi-model collaboration \cite{malach2017decoupling} and iterative strategies \cite{li2019dividemix}}).
Hybrid approaches are designed in practice.
For example, PeerLearn \cite{sun2021webly} develops a two-stage framework with peer models.
Each model chooses clean samples independently and feeds them to train the other model.
Different from the existing methods, we do not assume that samples with small losses or high confidence are clean.
Instead, we maintain class prototypes and filter out noise by comparing instances and prototypes in a non-linear metric.
Moreover, PeerLearn presumes that the percentage of noise is consistent across categories,
which contradicts our observation.

\subsection{Contrastive Representation Learning}

Contrastive learning methods can be roughly categorized as:
1) context-instance contrast, where the relationship of local parts with respect to global context is learned \cite{kim2018learning};
2) instance-wise contrast, where similar image pairs are pulled closer with dissimilar pairs pushed farther \cite{he2020momentum, chen2020simple}.
Prototypical contrastive learning (PCL) \cite{li2020prototypical} encourages each image embedding to be adjacent to its assigned cluster prototype.
However, their method is under an unsupervised setting where k-means clustering is used to generate prototypes.
Our model is supervised by both numerous-yet-noisy web labels and limited-yet-clean few-shot labels.
Besides, in PCL, batch embeddings in the current epoch are contrasted with the ``outdated" prototypes in the previous epoch.
FoPro keeps modifying prototypes smoothly all the time so that clean samples can be pinpointed by the latest features.

\begin{figure*}[ht]
\centering
\includegraphics[width=0.9\textwidth]{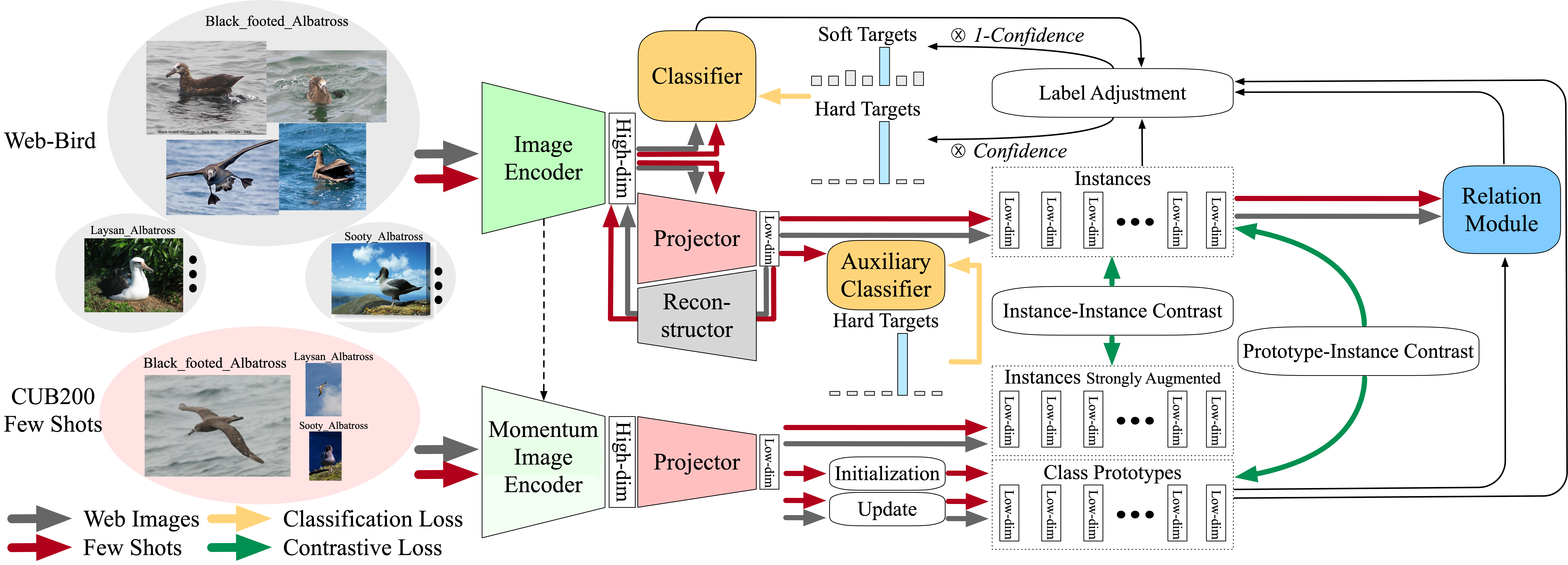} 
\caption{Overview of FoPro. The encoder, classifier, and projector are trained to produce discriminative embeddings. Class prototypes are first initialized by few shots and then polished with clean samples for contrastive learning to regularize cluster distribution. Instance-wise contrastive loss is optimized simultaneously. The relation module learns a distance metric between an instance and its assigned class prototype. Finally, we adjust web labels for confidence-weighted hybrid target learning.}
\label{fig:overview}
\end{figure*}

\section{Method}

In this section, a formal description of our few-shot WSL setting is presented, followed by the detailed explanation of FoPro.
Fig. \ref{fig:overview} illustrates the model architecture. 

\subsection{Problem Statement}

Existing WSL setting aims to train a deep model $\mathcal{F}(\theta_{e};\theta_{c})$ with the optimal parameters of encoder $\theta_{e}^{*}$ and classifier $\theta_{c}^{*}$ from the web dataset $\mathit{D}^{w}=\{(\mathbf{x}_i^{w}, y_i^{w})\}_{i=1}^{N^{w}}$.
Here, $\mathbf{x}_i^{w}$ denotes an image, $y_i^{w}\in\{1,...,C\}$ is its class label. The number of classes and images are $C$ and $N^{w}$, respectively.
Due to noise issues, $y_i^{w}$ might not equal to the ground-truth $y_i^{*}$.
If $y_i^{w}\neq y_i^{*}$ and $y_i^{*}\in\{1,...,C\}$, $(\mathbf{x}_i^{w}, y_i^{w})$ is viewed as an in-distribution (IND) sample with label-flipping error.
If $y_i^{w}\neq y_i^{*}$ and $y_i^{*}\notin\{1,...,C\}$, then $(\mathbf{x}_i^{w}, y_i^{w})$ is an out-of-distribution (OOD) sample.

We propose a new WSL setting that additional real-world images are available with verified labels:
$\mathit{D}^{t}=\{(\mathbf{x}_i^{t}, y_i^{t})\}_{i=1}^{N^{t}}$ and $y_i^{t}=y_i^{*}$.
The number of real-world samples is $N^{t}=K\cdot C$, where $K$ denotes $K$-shot per class.
Our FoPro aims to achieve two goals with $\mathit{D}^{w}$:
1) to learn generalizable representations from high-quality examples;
2) to correct IND samples and discard OOD samples.

\subsection{Model Architecture}

The main components of FoPro include siamese encoder backbones, a classifier, a projector, a reconstructor, an auxiliary classifier, and a relation module.

Our siamese encoder networks share the same architecture.
Enlighted by MoCo \cite{he2020momentum}, we update parameters of the first encoder $\theta_{e}^{1}$ by back-propagation and employ momentum update for the second encoder $\theta_{e}^{2}$:
\begin{equation}
    \theta_{e}^{2} = m_{e}\theta_{e}^{2}+(1-m_{e})\theta_{e}^{1},
\end{equation}
where $m_e$ is the momentum parameter.
The plain and momentum encoders respectively extract features $\mathbf{v}_i^{\{w;t\}}$ and $\mathbf{v}_{i}'^{\{w;t\}}\in{\rm I\!R}^{d_e}$ from inputs $\mathbf{x}_i^{\{w;t\}}$ and their augmented counterparts $\mathbf{x}_{i}'^{\{w;t\}}$.
Note that our encoder is structure-agnostic, and its choices are up to specific tasks.
All layers except the last fully connected (FC) layer are used.

A classifier is trained to map features $\mathbf{v}_i^{\{w;t\}}$ to the predicted probabilities $\mathbf{p}_i^{\{w;t\}}$ over $C$ classes.
It consists of one FC layer with softmax activation.

A projector distills discriminative contents from features $\mathbf{v}_i^{\{w;t\}}$ for low-dimensional embeddings $\mathbf{z}_i^{\{w;t\}}\in{\rm I\!R}^{d_p}$.
It is composed of two FC layers and one ReLU layer.
We follow \cite{chen2020simple, chen2020improved} to perform contrastive learning in the embedding space after projection.
$\ell_2$-normalization is involved for unit-sphere constraint on $\mathbf{z}_i^{\{w;t\}}$.

A reconstructor recovers $\tilde{\mathbf{v}}_i^{\{w;t\}}$ from $\mathbf{z}_i^{\{w;t\}}$,
where $\tilde{\mathbf{v}}_i^{\{w;t\}}$ should be as close as possible to $\mathbf{v}_i^{\{w;t\}}$. Symmetric structure is adopted for the projector and reconstructor.

An auxiliary classifier with one FC layer generates probabilities $\mathbf{q}_i^{t}$ over $C$ classes based on embeddings $\mathbf{z}_i^{t}$.

Our relation module compares each pair of one instance embedding $\mathbf{z}_i^{\{w;t\}}$ and one class prototype $\mathbf{c}_k\in{\rm I\!R}^{d_p}, k\in\{1,...,C\}$ for distance measurement.
Given the concatenated embeddings $[\mathbf{z}_i^{\{w;t\}}, \mathbf{c}_k]$,
it learns to score their closeness $r_{ik}\in{\rm I\!R}$ by two FC layers with one ReLU layer.

\subsection{Training Strategy}
FoPro employs a four-stage training strategy.

\subsubsection{Stage 1: Preparation} In this early stage, we warm up the system by learning common, regular patterns for the first $T_1$ epochs. As discovered by \cite{arpit2017closer}, easy examples are reliably learned with simple patterns before the model overfits noise.
We achieve this via training the encoder and classifier with cross-entropy loss.
\begin{equation}\label{eq:cls}
    \mathcal{L}_{i}^{cls}=-\log(\mathbf{p}_{i(y_i)}^{\{w;t\}}).
\end{equation}

Since $\mathbf{v}_i^{\{w;t\}}$ might contain redundant features that make outliers indistinguishable,
we set a projector to only keep principal components.
The previous method PCL stems from the analogy of autoencoder to PCA,
and learns projection by minimizing the reconstruction loss for the projector and reconstructor.
In preliminary experiments, however, we find that such optimization cannot give a good starting point for prototype initialization because $\mathbf{z}_i^{\{w;t\}}$ is not necessarily class-indicative.
Therefore, an auxiliary classifier is applied on $\mathbf{z}_i^{t}$ to bring back its representation capacity.
Only few shots are used here due to purity concerns.
\begin{equation}\label{eq:projection}
    \mathcal{L}_{i}^{prj}=\Vert\tilde{\mathbf{v}}_i^{\{w;t\}}-\mathbf{v}_i^{\{w;t\}}\Vert^{2}_{2}-\log(\mathbf{q}_{i(y_i)}^{t}).
\end{equation}

\subsubsection{Stage 2: Incubation} Clean few shots play an anchoring role in ``territory" enclosure in the embedding space.
Given extracted embeddings from the momentum encoder, we initialize prototypes by averaging few shots in each class.
\begin{equation}\label{eq:protoinit}
    \hat{\mathbf{c}}_k=\frac{1}{K}\sum_{y_i=k}{\mathbf{z}_i^{t}}, \mathbf{c}_k=\frac{\hat{\mathbf{c}}_k}{\Vert\hat{\mathbf{c}}_k\Vert_2}.
\end{equation}

In this stage, we begin to pull instances within one class towards the prototype for $T_2$ epochs.
Besides, instance-level discrimination is encouraged by contrastive losses \cite{chen2020simple} to improve separation across classes.
\begin{equation}\label{eq:prototype}
    \mathcal{L}_{i}^{pro}=-\log\frac{\exp((\mathbf{z}_i^{w;t}\cdot\mathbf{c}_{y_i}-\delta^{w;t})/\phi_{y_i})}{\sum_{k=1}^{C}\exp((\mathbf{z}_i^{w;t}\cdot\mathbf{c}_{k}-\delta^{w;t})/\phi_{k})},
\end{equation}
\begin{equation}\label{eq:instance}
    \mathcal{L}_{i}^{ins}=-\log \frac{\exp(\mathbf{z}_{i}^{w;t}\cdot\mathbf{z}_i'^{w;t}/\tau)}{\sum_{j=1}^{Q}\exp(\mathbf{z}_i^{w;t}\cdot\mathbf{z}_j'^{w;t}/\tau)},
\end{equation}
where $\delta^{w;t}$ refers to the margin coefficient, and $Q$ is the length of the memory bank for storing embeddings of visited instances.
Temperature coefficients can be fixed as $\tau$ or class-dependent as $\phi_{k}$.
We put constraints on learning representations with a high margin so that clean few shots gather around prototypes tightly,
ensuring better justification and interpretability.
Furthermore, to regularize the distribution of each class cluster, adjustable temperature coefficients \cite{li2020prototypical} are estimated based on concentration.
\begin{equation}\label{eq:concentrate}
    \phi_{k}=\frac{\sum_{y_{i}=k}\Vert\mathbf{z}_i^{w;t}-\mathbf{c}_{k}\Vert_2}{N_k^{w;t}\log(N_k^{w;t}+\alpha)},
\end{equation}
where $N_k^{w;t}$ denotes the number of web and few-shot instances of class $k$, and $\alpha$ is a smoother.
Embeddings of large, loose clusters will be penalized more to approach their prototypes, while those of small, tight clusters will be relaxed.

\subsubsection{Stage 3: Illumination} With parameters of the encoder and projector fixed, the relation module learns to score the compatibility between one instance and each prototype.
It sheds light on whether the given label of a web image is correct.
We select clean samples $\mathit{D}^{r}$ for training the relation module.
\begin{equation}\label{eq:cleanset}
    \mathit{D}^{r} = \mathit{D}^{t} \cup \{(\mathbf{x}_i^{w}, y_i^{w})|\sum_{j=1}^{C}\vert(\mathbf{z}_{i}^{w}-\mathbf{c}_{y_i})\cdot\mathbf{c}_{j}\vert\leq\sigma\},
\end{equation}
where $\sigma$ is a threshold between 0 and 1. Such criterion comprehensively considers both the cosine distance between instance and prototypes, and the distance among prototypes.
Then, the relation module is trained for $T_3$ epochs by:
\begin{equation}\label{eq:rel}
    \mathcal{L}_{i}^{rel}=-\log\frac{\exp(r_{i y_i})}{\sum_{k=1}^{C}\exp(r_{ik})}.
\end{equation}

\subsubsection{Stage 4: Verification} Armed with ``pretrained" model, we start label correction, OOD removal, prototype update, and continue noise-robust learning for $T_4$ epochs.
Three sources of prior knowledge are incorporated for cleaning:
1) self-prediction;
2) instance-prototype similarity;
3) relation score.
Rules for adjusting labels are detailed below:
\begin{equation}\label{eq:gtscore}
    \mathbf{s}_i^{w}=\beta\mathbf{p}_i^{w} + (1-\beta)[\mathbf{c}_{1},...,\mathbf{c}_{C}]^{T}\cdot\mathbf{z}_i^{w}
\end{equation}
\begin{equation}\label{eq:update}
    \hat{y}_{i}^{w}=\left\{
\begin{array}{lcl}
y_{i}^{w}    &  & {\textrm{if}\ r_{i y_{i}}>\gamma,} \\
\arg\max_{k}\mathbf{s}_{i(k)}^{w}  &  & {\textrm{else if} \max_{k}\mathbf{s}_{i(k)}^{w}>\gamma,} \\
y_{i}^{w}    &  & {\textrm{else if}\ \mathbf{s}_{i(y_i)}^{w}>1/C,} \\
\textrm{Null}\ (OOD)   &  & {\textrm{otherwise,}}
\end{array} \right.
\end{equation}
where $\gamma$ is a threshold between 0 and 1.
Since fine-grained categories share highly similar visual patterns,
the relation module is only used for positive verification of the initial web label.
Besides, we introduce an alternative confidence measure from self-prediction and similarity for label reassignment.
When the first two conditions are not satisfied, an image will be kept as hard example if its confidence is above average.
Otherwise, it is discarded as OOD.
Note that the basic control flow above is inspired by MoPro.
We further improve it with the proposed relation module (Eq.\ref{eq:update} cond. 1) to better evaluate the compatibility between instances and class prototypes,
and thereafter enable accurate label-flipping-error correction and OOD removal without ignoring hard examples by mistake (Eq.\ref{eq:update} conds. 2--4).

After label adjustment, we exploit the predicted probabilities as pseudo-labels for self-training \cite{tanaka2018joint, han2019deep}.
Such soft targets can be viewed as a regularizer on the classifier like label smoothing \cite{muller2019does} and self-knowledge distillation \cite{hinton2015distilling}.
Instead of using a fixed coefficient, we follow \cite{yang2020webly} to leverage confidence on the corrected label for weighting soft and hard targets.
\begin{eqnarray}\label{eq:cls2}
    \mathcal{L}_{i}^{cls}=&-\log(\mathbf{p}_{i(y_i)}^{t})-\mathbf{s}_{i(\hat{y}_{i})}^{w}\log(\mathbf{p}_{i(\hat{y}_i)}^{w})\nonumber\\
    &-(1-\mathbf{s}_{i(\hat{y}_{i})}^{w})\sum_{k=1}^{C}\mathbf{p}_{i(k)}^{w}\log\mathbf{p}_{i(k)}^{w}.
\end{eqnarray}

With the label-flipping errors and OOD reduced, class prototypes are updated by embeddings of the remaining clean examples from the momentum encoder.
Exponential moving average \cite{li2020mopro} is used for two reasons:
1) initialization by few shots remains to exert a profound anchoring effect.
2) smoothed transition is achieved to stabilize contrastive learning.
For class $k$, web images with $\hat{y}_{i}^{w}=k$ and few shot images with $y_{i}^{t}=k$ are involved:
\begin{equation}\label{eq:protoupdate}
    \hat{\mathbf{c}}_k=m_{p}{\mathbf{c}}_k+(1-m_{p})\mathbf{z}_i^{w;t}, \mathbf{c}_k=\frac{\hat{\mathbf{c}}_k}{\Vert\hat{\mathbf{c}}_k\Vert_2}.
\end{equation}

Note that reliable samples, which are selected by Eq. \ref{eq:update} per batch, also participate in training the relation module.
The criterion by Eq. \ref{eq:cleanset} is only used in stage 3.

\section{Experiments}

We train FoPro on web datasets and evaluate it on real-world testing sets.
FoPro boosts $K$-shot performance and reaches the SOTA.
Ablation study validates the relation module.

\subsection{Datasets}

\begin{table}[htbp]
\centering\fontsize{9pt}{10pt}\selectfont{
\begin{tabular}{cllll}
\hline
\multicolumn{2}{l}{Web Dataset} & \# Img. & \# Cls. & Real-World \\ \hline
\multirow{3}{*}{\begin{tabular}[c]{@{}c@{}}Web-\\ FG496\end{tabular}} & Bird & 18k & 200 & CUB200-2011 \\
 & Air & 13k & 100 & FGVC-Aircraft \\
 & Car & 21k & 196 & Stanford Car \\ \hline
\multirow{2}{*}{\begin{tabular}[c]{@{}c@{}}Web-\\ Vision1k\end{tabular}} & \begin{tabular}[c]{@{}l@{}}Web-\\ Vision1k\end{tabular} & 2.44M & 1k & ImageNet1k \\
 & Google500 & 0.61M & 500 & ImageNet500 \\ \hline
\end{tabular}
}
\caption{Statistics of web datasets.}
\label{tab:dataset}
\end{table}

\paragraph{WebFG496} \cite{sun2021webly} contains three fine-grained datasets sourced from Bing.
The testing sets of CUB200-2011 \cite{wah2011caltech}, FGVC-Aircraft \cite{maji2013fine}, and Stanford Car \cite{krause20133d} are used.

\paragraph{WebVision1k} \cite{li2017webvision} is collected from Google and Flickr.
The validation set of ImageNet1k \cite{deng2009imagenet} is used.
Besides, we also use \textbf{Google500} \cite{yang2020webly} where 500 out of 1k categories are randomly sampled with images only from Google (see Table \ref{tab:dataset}).

We randomly sample $K$ shots per class from the training sets of real-world datasets.
Classification accuracy (\%) is adopted as the evaluation metric for all experiments.

\subsection{Implementation Details}

\paragraph{WebFG496} The B-CNN \cite{lin2015bilinear} (VGG-16 \cite{simonyan2014very}) is used as encoder.
We refer to \cite{sun2021webly} for the training settings:
optimizer is Adam with weight decay of $1\times 10^{-8}$;
batch size is 64;
the learning rate is $1\times 10^{-4}$ and decays to 0 by cosine schedule;
a warm-up policy increases the learning rate linearly for 5 epochs with the frozen encoders.

\paragraph{WebVision1k} The ResNet-50 (R50) \cite{he2016deep} is used as encoder.
We refer to \cite{yang2020webly} for the training settings:
batch size is 256;
optimizer is SGD with the momentum of 0.9 and weight decay of $1\times 10^{-4}$;
the learning rate is 0.01 and decays to 0 by cosine schedule.

We refer to MoPro to set $m_e=0.999$, $m_p=0.999$, $d_p=128$, and $Q=8192$.
In view of the dataset scale, we set $T_1=20$, $T_2=5$, $T_3=20$, $T_4=175$ for WebFG496 and set $T_1=15$, $T_2=5$, $T_3=10$, $T_4=30$ for WebVision1k/Google500.
Preliminary experiments on WebFG496 show that $\gamma=0.6$ and $\beta=0.5$ work better than $\gamma=0.2$ and $\beta=0, 0.25, 0.75, 1$.
A lower $\gamma$ means a more relaxed criterion on clean sample selection,
which might bring in noise and cause performance drop.
The balanced combination of self-prediction and similarity terms performs more robust to noise than the biased cases. Other hyper-parameters are empirically set as: $\delta^{w}=0$, $\delta^{t}=0.5$, $\tau=0.1$, $\alpha=10$, $\sigma=20$.
Their optimal values require meticulous fine-tuning, which is beyond consideration of the present study.

Data augmentation includes random cropping and horizontal flipping.
Strong augmentation on the inputs to the momentum encoder \cite{he2020momentum} additionally uses color jittering and blurring.
Since birds might only differ in color, random rotation in 45 degrees is used instead.
Experiments are conducted on a CentOS 7 workstation with an Intel 8255C CPU, 377 GB Mem, and 8 NVIDIA V100 GPUs.

\subsection{Results}

\begin{table}[htbp]
\begin{threeparttable}
\fontsize{9pt}{10pt}\selectfont{
\begin{tabular}{llllll}
\hline
\multirow{2}{*}{Method} & Back- & \multicolumn{4}{c}{WebFG496} \\
 & bone & Bird & Air & Car & Avg. \\ \hline
Vanilla & R50 & 64.43 & 60.79 & 60.64 & 61.95 \\
MoPro$^\dagger$ & R50 & 71.16 & 76.85 & 79.68 & 75.90 \\ \hline
SCC$^\dagger$ & R50-D  & 61.10 & 74.92 & 83.49 & 73.17 \\ \hline
Vanilla & B-CNN & 66.56 & 64.33 & 67.42 & 66.10 \\
Decouple & B-CNN & 70.56 & 75.97 & 75.00 & 73.84 \\
CoTeach & B-CNN & 73.85 & 72.76 & 73.10 & 73.24 \\
PeerLearn & B-CNN & 76.48 & 74.38 & 78.52 & 76.46 \\
PeerLearn$^\dagger$ & B-CNN & 76.57 & 74.35 & 78.26 & 76.39 \\ \hline
FoPro($K$=0)  & B-CNN & 77.79 & 79.37 & 86.99 & 81.38 \\
FoPro($K$=1)  & B-CNN & {78.07} & {79.87} & {88.01} & {82.03} \\
FoPro($K$=16)  & B-CNN & \textbf{85.54} & \textbf{86.40} & \textbf{91.51} & \textbf{87.81} \\ \hline
\end{tabular}
}
\begin{tablenotes}
\footnotesize
\item[$\dagger$] Results are reproduced by ourselves with the official codes.
\end{tablenotes}
\caption{The SOTA results on fine-grained datasets.}
\label{tab:comp1}
\end{threeparttable}
\end{table}

\paragraph{Baselines} Our FoPro is compared with vanilla backbones and the SOTA methods including SCC, MoPro, Decouple \cite{malach2017decoupling}, CoTeach \cite{han2018co}, PeerLearn, MentorNet \cite{jiang2018mentornet}, CurriculumNet \cite{guo2018curriculumnet}, and CleanNet \cite{lee2018cleannet}.
Results of the SOTA methods that are trained and evaluated on the same datasets are directly quoted here.
We also reproduce three closely-related methods of SCC, MoPro, and PeerLearn under $K$-shot settings with the officially released codes.
Their default hyper-parameters are employed if the same web datasets are engaged. Otherwise, they are set the same as ours.
Additionally, we modify the proposed method only to exhibit its applicability for 0-shot without specific design.
In that case, web images with predicted probability of the target class over $\gamma$ are used to train the auxiliary classifier.
In view of the dataset scale, prototypes are initialized by randomly sampled 16 and 50 web images per class from WebFG496 and WebVision1k/Google500, respectively.

\begin{table}[htbp]
\begin{threeparttable}
\fontsize{9pt}{10pt}\selectfont{
\begin{tabular}{llllll}
\hline
\multirow{2}{*}{Method$^\dagger$} & Back- & \multicolumn{2}{c}{ImageNet1k} & \multicolumn{2}{c}{ImageNet500} \\
 & bone & Top 1 & Top 5 & Top 1 & Top 5 \\
 \hline
\multirow{2}{*}{MentorNet} & \multirow{2}{*}{\begin{tabular}[c]{@{}l@{}}Inception\\ResNetV2\end{tabular}} & \multirow{2}{*}{64.20} & \multirow{2}{*}{84.80} & \multirow{2}{*}{--} & \multirow{2}{*}{--} \\ \\
\multirow{2}{*}{\begin{tabular}[c]{@{}l@{}}Curriculum-\\Net\end{tabular}} & \multirow{2}{*}{\begin{tabular}[c]{@{}l@{}}Inception\\V2\end{tabular}} & \multirow{2}{*}{64.80} & \multirow{2}{*}{83.40} & \multirow{2}{*}{--} & \multirow{2}{*}{--} \\ \\ \hline
Vanilla & R50-D & 67.23 & 84.09 & -- & -- \\
SCC & R50-D & 67.93 & 84.77 & 68.84 & 84.62 \\
SCC$^\dagger$ & R50-D & 67.57 & 85.74 & 64.40 & 81.56 \\ \hline
Vanilla & R50 & 65.70 & 85.10 & 61.54 & 78.89 \\ 
CoTeach & R50 & -- & -- & 62.18 & 80.98 \\
CleanNet & R50 & 63.42 & 84.59 & -- & -- \\ 
MoPro & R50 & 67.80 & 87.00 & -- & -- \\
MoPro$^\dagger$ & R50 & 66.05 & 85.66 & 58.68 & 78.39 \\
PeerLearn$^\dagger$ & R50 & 52.57 & 73.35 & 42.04 & 61.71 \\ \hline
FoPro($K$=0) & R50 & 67.03 & 85.57 & 68.59  & 86.03 \\
FoPro($K$=1) & R50 & 67.55 & 86.31 & 69.11 & 86.19 \\
FoPro($K$=16) & R50 & \textbf{68.83} & \textbf{87.83}  & \textbf{72.02} & \textbf{89.38} \\
\hline
\end{tabular}
}
\begin{tablenotes}
\footnotesize
\item[$\dagger$] Results are reproduced by ourselves with the official codes.
\end{tablenotes}
\end{threeparttable}
\caption{The SOTA results on large-scale datasets.}
\label{tab:comp2}
\end{table}

Table \ref{tab:comp1} confirms the superiority of the proposed method on fine-grained datasets even under 0-shot.
FoPro boosts the accuracy of vanilla backbones more significantly than the SOTA methods with respect to their backbones.

FoPro reaches the optimal performance on large-scale datasets with $K$=16 (see Table \ref{tab:comp2}). The vanilla R50-D \cite{he2019bag} performs better than R50. Although FoPro is preceded by SCC and MoPro at first (0-shot), it rises steadily after exploiting a few real-world examples efficiently.

\begin{figure}[htbp]
\centering
\includegraphics[width=0.99\columnwidth]{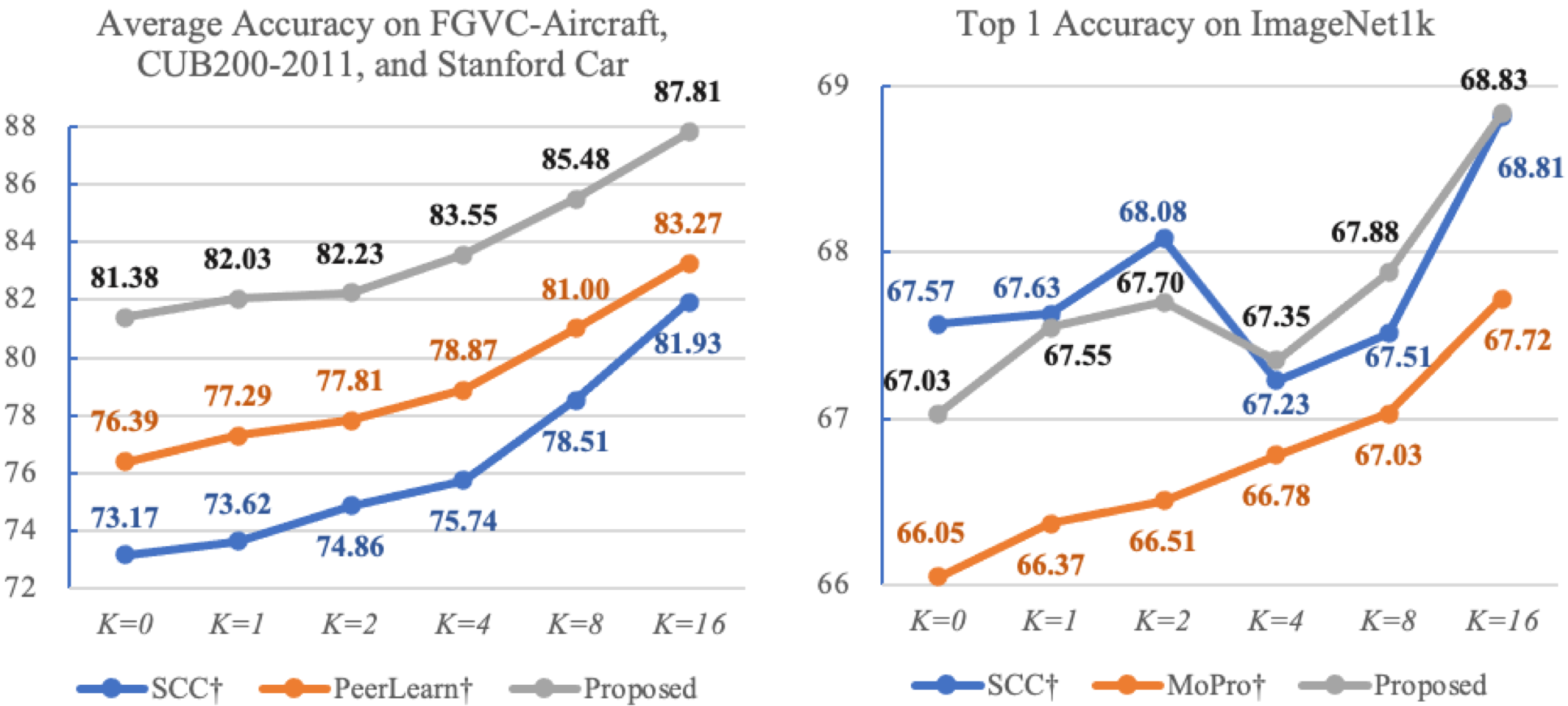}
\caption{The SOTA results under $K$-shot settings.}
\label{fig:compk}
\end{figure}

\begin{figure*}[htbp]
\centering
\includegraphics[width=0.99\textwidth]{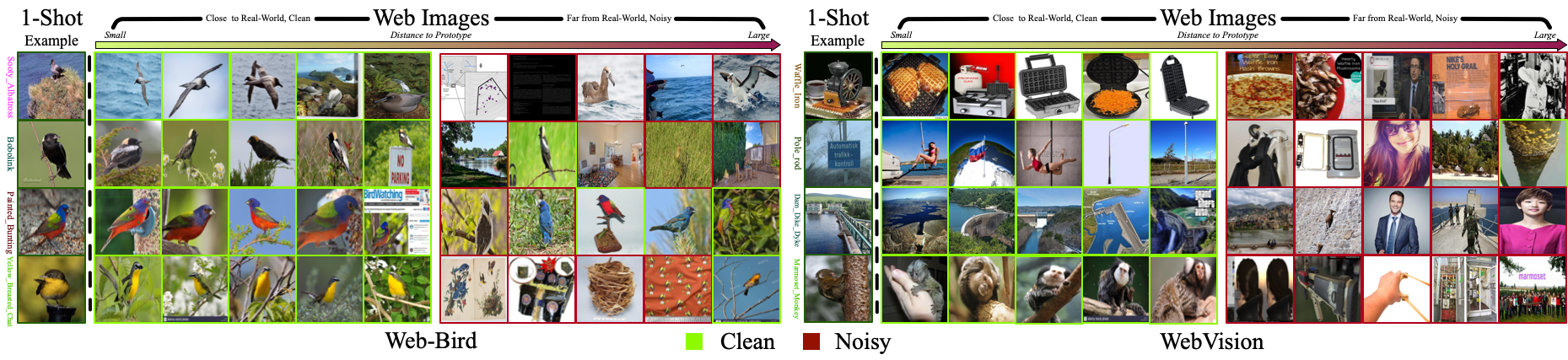}
\caption{One-shot real-world examples and the web images sorted by their distance to class prototypes. Best viewed magnified.}
\label{fig:visual}
\end{figure*}

Under the degeneration circumstance ($K$=0),
FoPro outperforms the SOTA methods on WebFG496.
The reason why FoPro ($K$=0) degrades slightly on ImageNet1k/500 lies in the high percentage of noises in WebVision1k/Google500.
In that case,
prototypes ($K$=0) are initialized and polished solely by noisy web examples without intervention from clean shots,
which may not be class-representative.
With few real-world examples ($K>0$) involved,
FoPro regains its advantage over the SOTA methods.

\begin{table}[htbp]
\begin{threeparttable}
\fontsize{9pt}{10pt}\selectfont{
\begin{tabular}{lllllll}
\hline
\multirow{2}{*}{$K$} & \multicolumn{2}{c}{WebFG496 Avg.} & \multicolumn{2}{c}{ImageNet1k} & \multicolumn{2}{c}{ImageNet500} \\
 & Top 1 & Gap & Top 1 & Gap & Top 1 & Gap \\ \hline
0 & 81.38 & -- & 67.03 & 5.57 & 68.59 & 3.85 \\ \hline
1 & +0.65 & -- & +0.52 & 5.22 & +0.52 & 3.63 \\
2 & +0.85 & -- & +0.67 & 5.20 & +1.35 & 3.29 \\
4 & +2.17 & -- & +0.32 & 4.60 & +1.50 & 2.91 \\
8 & +4.10 & -- & +0.85 & 4.64 & +2.06 & 2.90 \\
16 & +6.43 & -- & +1.80 & 3.91 & +3.43 & 2.19 \\ \hline
16 & 87.81 & -- & 68.83 & -- & 72.02 & -- \\
Ref. & 87.16$^{\dagger}$ & -- & 76.15$^{\ddagger}$ & -- & 76.22$^{\ddagger}$ & -- \\
\hline
\end{tabular}
}
\begin{tablenotes}
\footnotesize
\item[$\dagger$] Official results of the B-CNN trained on FGVC-Aircraft, CUB200-2011, and Stanford Car are averaged.
\item[$^{\ddagger}$] Official results of the R50 trained on ImageNet1k by PyTorch are quoted respectively for 500 and 1k classes.
\end{tablenotes}
\caption{FoPro gains of $K$-shot over 0-shot. Gap refers to the differences between web and real-world testing results.}
\label{tab:potential}
\end{threeparttable}
\end{table}

\paragraph{Effect of Few-Shots} We explore the potential of FoPro by varying the number of real-world examples per class from 1 to 16.
As shown in Fig. \ref{fig:compk}, FoPro achieves consistent performance growth with $K$ on fine-grained datasets.
It surpasses SCC and PeerLearn by a large margin.
On ImageNet1k,
the abnormal case of $K$=4 is mainly due to sampling jittering.
Since ImageNet contains many unreal images,
it could not eliminate the possibility of sampling atypical images of certain classes.
However,
as $K$ increases,
FoPro starts to take the lead.
We believe more few shots directly refine the estimated prototypes for better representation.
Clean samples can be more appropriately selected to promote discriminative feature learning.
With amendment on cluster formation, FoPro also enjoys a higher level of interpretability in class centers with competitive performance.

Table \ref{tab:potential} reports the performance gap between WebVision1k/ImageNet1k and Google500/ImageNet500 when the testing sets of web domain are available.
In line with Fig. \ref{fig:proposed}(b), the reduced gap reflects that we bridge the noisy web domain and real-world domain with limited $K$ shots.
FoPro approaches reference benchmarks that are trained on real-world datasets, corroborating its practical value that much labor of data collection and annotation can be saved.

\paragraph{Effect of Relation Module} In Table \ref{tab:ablation}, we study the effect of relation module for clean example selection.
We remove FC layers and directly compare an instance and each prototype using cosine similarity.
Results confirm that the proposed relation module discovers clean examples more precisely than the pre-defined similarity metric.
By using a non-linear metric, we do not assume that element-wise comparison could solely separate matching or mismatching pairs.
Besides, such a learnable metric is not sensitive to input variation and behaves better on noisy samples.

\begin{table}[htbp]
\centering\fontsize{9pt}{10pt}\selectfont{
\begin{tabular}{llll}
\hline
$K$=1 & WebFG496 Avg. & ImageNet1k & ImageNet500 \\ \hline
w/o RM & 81.59 & 65.22 & 64.69 \\
w RM & 82.03 & 67.55 & 69.11 \\\hline
\end{tabular}
\caption{Ablation Study on the Relation Module (RM).}
\label{tab:ablation}
}
\end{table}

\paragraph{Visualization} In Fig. \ref{fig:proposed}(a), we visualize the low-dimensional embeddings with t-SNE for the randomly chosen 10 categories in WebVision as a demonstration.
For convenience,
all 16 real-world examples in each class are averaged and displayed as one few-shot example.
Differences in the cluster distribution (from $K$=1 to $K$=16) are highlighted to show that:
1) the distance between each prototype and the few-shot example becomes shorter;
2) the density of class clusters is improved.
From Fig. \ref{fig:visual}, we conclude the following insights:
1) Web images close to the estimated prototypes are clean and similar to real-world photos with limited post-processing.
Our FoPro learns to sort out noise in web data for robust representation learning.
2) The proposed method generalizes across various domains such as product close-up, computer graphics, and screenshots.
3) Intra-class diversity (\eg{wing postures of the sooty albatross}),
uncaptured salient parts (\eg{the yellowish patch on the back head of the bobolink}),
and editing of tone curve (\eg{colored body of the painted bunting and yellow-breasted chat}) are the possible reasons why hard examples of 1-shot and clean web images still locate away from prototypes.

\section{Conclusion}
This paper introduces a new setting for webly-supervised learning,
which optimizes the learning system with a large quantity of noisy web images and a few real-world images.
Under this setting, we propose a few-shot guided prototypical representation learning method called FoPro,
which simultaneously tackles noise and dataset bias in a cost-efficient manner.
It is characterized by the guidance from a few real-world domain images for learning noise-robust and generalizable representations from web data.
Experimental results demonstrate that our method can effectively utilize few-shot images to improve the performance of WSL on real-world benchmarks.
Future work includes investigation of side information from web datasets (\eg{captions, website titles, and user comments}) and application extension to weakly-supervised object detection and segmentation.


\begin{thebibliography}{58}
\providecommand{\natexlab}[1]{#1}

\bibitem[{Arpit et~al.(2017)Arpit, Jastrzebski, Ballas, Krueger, Bengio,
  Kanwal, Maharaj, Fischer, Courville, Bengio et~al.}]{arpit2017closer}
Arpit, D.; Jastrzebski, S.; Ballas, N.; Krueger, D.; Bengio, E.; Kanwal, M.~S.;
  Maharaj, T.; Fischer, A.; Courville, A.; Bengio, Y.; et~al. 2017.
\newblock A closer look at memorization in deep networks.
\newblock In \emph{International Conference on Machine Learning}, 233--242.
  PMLR.

\bibitem[{Bergamo and Torresani(2010)}]{bergamo2010exploiting}
Bergamo, A.; and Torresani, L. 2010.
\newblock Exploiting weakly-labeled web images to improve object
  classification: a domain adaptation approach.
\newblock \emph{Advances in Neural Information Processing Systems}, 23.

\bibitem[{Chen et~al.(2020{\natexlab{a}})Chen, Kornblith, Norouzi, and
  Hinton}]{chen2020simple}
Chen, T.; Kornblith, S.; Norouzi, M.; and Hinton, G. 2020{\natexlab{a}}.
\newblock A simple framework for contrastive learning of visual
  representations.
\newblock In \emph{International Conference on Machine Learning}, 1597--1607.
  PMLR.

\bibitem[{Chen et~al.(2020{\natexlab{b}})Chen, Fan, Girshick, and
  He}]{chen2020improved}
Chen, X.; Fan, H.; Girshick, R.; and He, K. 2020{\natexlab{b}}.
\newblock Improved baselines with momentum contrastive learning.
\newblock \emph{arXiv preprint arXiv:2003.04297}.

\bibitem[{Chen and Gupta(2015)}]{chen2015webly}
Chen, X.; and Gupta, A. 2015.
\newblock Webly supervised learning of convolutional networks.
\newblock In \emph{Proceedings of the IEEE international conference on computer
  vision}, 1431--1439.

\bibitem[{Cheng et~al.(2020)Cheng, Zhou, Zhao, Li, Shang, Zheng, Pan, and
  Xu}]{cheng2020weakly}
Cheng, L.; Zhou, X.; Zhao, L.; Li, D.; Shang, H.; Zheng, Y.; Pan, P.; and Xu,
  Y. 2020.
\newblock Weakly supervised learning with side information for noisy labeled
  images.
\newblock In \emph{European Conference on Computer Vision}, 306--321. Springer.

\bibitem[{Deng et~al.(2009)Deng, Dong, Socher, Li, Li, and
  Fei-Fei}]{deng2009imagenet}
Deng, J.; Dong, W.; Socher, R.; Li, L.-J.; Li, K.; and Fei-Fei, L. 2009.
\newblock ImageNet: A large-scale hierarchical image database.
\newblock In \emph{Proceedings of the IEEE/CVF International Conference on
  Computer Vision}, 248--255. IEEE.

\bibitem[{Divvala, Farhadi, and Guestrin(2014)}]{divvala2014learning}
Divvala, S.~K.; Farhadi, A.; and Guestrin, C. 2014.
\newblock Learning everything about anything: Webly-supervised visual concept
  learning.
\newblock In \emph{Proceedings of the IEEE Conference on Computer Vision and
  Pattern Recognition}, 3270--3277.

\bibitem[{Ghosh, Kumar, and Sastry(2017)}]{ghosh2017robust}
Ghosh, A.; Kumar, H.; and Sastry, P.~S. 2017.
\newblock Robust loss functions under label noise for deep neural networks.
\newblock In \emph{Proceedings of the AAAI Conference on Artificial
  Intelligence}, volume~31.

\bibitem[{Guo et~al.(2018)Guo, Huang, Zhang, Zhuang, Dong, Scott, and
  Huang}]{guo2018curriculumnet}
Guo, S.; Huang, W.; Zhang, H.; Zhuang, C.; Dong, D.; Scott, M.~R.; and Huang,
  D. 2018.
\newblock CurriculumNet: Weakly supervised learning from large-scale web
  images.
\newblock In \emph{European Conference on Computer Vision}, 135--150.

\bibitem[{Han et~al.(2018)Han, Yao, Yu, Niu, Xu, Hu, Tsang, and
  Sugiyama}]{han2018co}
Han, B.; Yao, Q.; Yu, X.; Niu, G.; Xu, M.; Hu, W.; Tsang, I.; and Sugiyama, M.
  2018.
\newblock Co-teaching: Robust Training of Deep Neural Networks with Extremely
  Noisy Labels.
\newblock \emph{Advances in Neural Information Processing Systems}, 31.

\bibitem[{Han, Luo, and Wang(2019)}]{han2019deep}
Han, J.; Luo, P.; and Wang, X. 2019.
\newblock Deep self-learning from noisy labels.
\newblock In \emph{Proceedings of the IEEE/CVF Conference on Computer Vision
  and Pattern Recognition}, 5138--5147.

\bibitem[{He et~al.(2020)He, Fan, Wu, Xie, and Girshick}]{he2020momentum}
He, K.; Fan, H.; Wu, Y.; Xie, S.; and Girshick, R. 2020.
\newblock Momentum contrast for unsupervised visual representation learning.
\newblock In \emph{Proceedings of the IEEE/CVF Conference on Computer Vision
  and Pattern Recognition}, 9729--9738.

\bibitem[{He et~al.(2016)He, Zhang, Ren, and Sun}]{he2016deep}
He, K.; Zhang, X.; Ren, S.; and Sun, J. 2016.
\newblock Deep residual learning for image recognition.
\newblock In \emph{Proceedings of the IEEE/CVF International Conference on
  Computer Vision}, 770--778.

\bibitem[{He et~al.(2019)He, Zhang, Zhang, Zhang, Xie, and Li}]{he2019bag}
He, T.; Zhang, Z.; Zhang, H.; Zhang, Z.; Xie, J.; and Li, M. 2019.
\newblock Bag of tricks for image classification with convolutional neural
  networks.
\newblock In \emph{Proceedings of the IEEE/CVF Conference on Computer Vision
  and Pattern Recognition}, 558--567.

\bibitem[{Hinton et~al.(2015)Hinton, Vinyals, Dean
  et~al.}]{hinton2015distilling}
Hinton, G.; Vinyals, O.; Dean, J.; et~al. 2015.
\newblock Distilling the knowledge in a neural network.
\newblock \emph{arXiv preprint arXiv:1503.02531}, 2(7).

\bibitem[{Jiang et~al.(2018)Jiang, Zhou, Leung, Li, and
  Fei-Fei}]{jiang2018mentornet}
Jiang, L.; Zhou, Z.; Leung, T.; Li, L.-J.; and Fei-Fei, L. 2018.
\newblock MentorNet: Learning data-driven curriculum for very deep neural
  networks on corrupted labels.
\newblock In \emph{International Conference on Machine Learning}, 2304--2313.
  PMLR.

\bibitem[{Jin, Ortiz~Segovia, and Susstrunk(2017)}]{jin2017webly}
Jin, B.; Ortiz~Segovia, M.~V.; and Susstrunk, S. 2017.
\newblock Webly supervised semantic segmentation.
\newblock In \emph{Proceedings of the IEEE/CVF Conference on Computer Vision
  and Pattern Recognition}, 3626--3635.

\bibitem[{Kaur, Sikka, and Divakaran(2017)}]{kaur2017combining}
Kaur, P.; Sikka, K.; and Divakaran, A. 2017.
\newblock Combining weakly and webly supervised learning for classifying food
  images.
\newblock \emph{arXiv preprint arXiv:1712.08730}.

\bibitem[{Kim et~al.(2018)Kim, Cho, Yoo, and Kweon}]{kim2018learning}
Kim, D.; Cho, D.; Yoo, D.; and Kweon, I.~S. 2018.
\newblock Learning image representations by completing damaged jigsaw puzzles.
\newblock In \emph{IEEE Winter Conference on Applications of Computer Vision},
  793--802. IEEE.

\bibitem[{Kolesnikov et~al.(2019)Kolesnikov, Beyer, Zhai, Puigcerver, Yung,
  Gelly, and Houlsby}]{kolesnikov2019large}
Kolesnikov, A.; Beyer, L.; Zhai, X.; Puigcerver, J.; Yung, J.; Gelly, S.; and
  Houlsby, N. 2019.
\newblock Large scale learning of general visual representations for transfer.
\newblock \emph{arXiv preprint arXiv:1912.11370}, 2(8).

\bibitem[{Krause et~al.(2016)Krause, Sapp, Howard, Zhou, Toshev, Duerig,
  Philbin, and Fei-Fei}]{krause2016unreasonable}
Krause, J.; Sapp, B.; Howard, A.; Zhou, H.; Toshev, A.; Duerig, T.; Philbin,
  J.; and Fei-Fei, L. 2016.
\newblock The unreasonable effectiveness of noisy data for fine-grained
  recognition.
\newblock In \emph{European Conference on Computer Vision}, 301--320. Springer.

\bibitem[{Krause et~al.(2013)Krause, Stark, Deng, and Fei-Fei}]{krause20133d}
Krause, J.; Stark, M.; Deng, J.; and Fei-Fei, L. 2013.
\newblock 3d object representations for fine-grained categorization.
\newblock In \emph{Proceedings of the IEEE/CVF International Conference on
  Computer Vision Workshops}, 554--561.

\bibitem[{Lee et~al.(2018)Lee, He, Zhang, and Yang}]{lee2018cleannet}
Lee, K.-H.; He, X.; Zhang, L.; and Yang, L. 2018.
\newblock CleanNet: Transfer Learning for Scalable Image Classifier Training
  with Label Noise.
\newblock In \emph{Proceedings of the IEEE/CVF Conference on Computer Vision
  and Pattern Recognition}, 5447--5456.

\bibitem[{Li, Socher, and Hoi(2019)}]{li2019dividemix}
Li, J.; Socher, R.; and Hoi, S.~C. 2019.
\newblock DivideMix: Learning with Noisy Labels as Semi-supervised Learning.
\newblock In \emph{International Conference on Learning Representations}.

\bibitem[{Li, Xiong, and Hoi(2020)}]{li2020mopro}
Li, J.; Xiong, C.; and Hoi, S. 2020.
\newblock MoPro: Webly Supervised Learning with Momentum Prototypes.
\newblock In \emph{International Conference on Learning Representations}.

\bibitem[{Li et~al.(2020)Li, Zhou, Xiong, and Hoi}]{li2020prototypical}
Li, J.; Zhou, P.; Xiong, C.; and Hoi, S. 2020.
\newblock Prototypical Contrastive Learning of Unsupervised Representations.
\newblock In \emph{International Conference on Learning Representations}.

\bibitem[{Li et~al.(2017)Li, Wang, Li, Agustsson, and
  Van~Gool}]{li2017webvision}
Li, W.; Wang, L.; Li, W.; Agustsson, E.; and Van~Gool, L. 2017.
\newblock WebVision database: Visual learning and understanding from web data.
\newblock \emph{arXiv preprint arXiv:1708.02862}.

\bibitem[{Lin, RoyChowdhury, and Maji(2015)}]{lin2015bilinear}
Lin, T.-Y.; RoyChowdhury, A.; and Maji, S. 2015.
\newblock Bilinear CNN models for fine-grained visual recognition.
\newblock In \emph{Proceedings of the IEEE/CVF International Conference on
  Computer Vision}, 1449--1457.

\bibitem[{Liu et~al.(2021)Liu, Zhang, Lu, and Tang}]{liu2021exploiting}
Liu, H.; Zhang, H.; Lu, J.; and Tang, Z. 2021.
\newblock Exploiting Web Images for Fine-Grained Visual Recognition via Dynamic
  Loss Correction and Global Sample Selection.
\newblock \emph{IEEE Transactions on Multimedia}, 24: 1105--1115.

\bibitem[{Maji et~al.(2013)Maji, Rahtu, Kannala, Blaschko, and
  Vedaldi}]{maji2013fine}
Maji, S.; Rahtu, E.; Kannala, J.; Blaschko, M.; and Vedaldi, A. 2013.
\newblock Fine-grained visual classification of aircraft.
\newblock \emph{arXiv preprint arXiv:1306.5151}.

\bibitem[{Malach and Shalev-Shwartz(2017)}]{malach2017decoupling}
Malach, E.; and Shalev-Shwartz, S. 2017.
\newblock Decoupling" when to update" from" how to update".
\newblock \emph{Advances in Neural Information Processing Systems}, 30.

\bibitem[{M{\"u}ller, Kornblith, and Hinton(2019)}]{muller2019does}
M{\"u}ller, R.; Kornblith, S.; and Hinton, G.~E. 2019.
\newblock When does label smoothing help?
\newblock \emph{Advances in Neural Information Processing Systems}, 32.

\bibitem[{Niu, Li, and Xu(2015)}]{niu2015visual}
Niu, L.; Li, W.; and Xu, D. 2015.
\newblock Visual recognition by learning from web data: A weakly supervised
  domain generalization approach.
\newblock In \emph{Proceedings of the IEEE Conference on Computer Vision and
  Pattern Recognition}, 2774--2783.

\bibitem[{Pereyra et~al.(2017)Pereyra, Tucker, Chorowski, Kaiser, and
  Hinton}]{pereyra2017regularizing}
Pereyra, G.; Tucker, G.; Chorowski, J.; Kaiser, {\L}.; and Hinton, G. 2017.
\newblock Regularizing neural networks by penalizing confident output
  distributions.
\newblock \emph{arXiv preprint arXiv:1701.06548}.

\bibitem[{Reed et~al.(2015)Reed, Lee, Anguelov, Szegedy, Erhan, and
  Rabinovich}]{reed2015training}
Reed, S.~E.; Lee, H.; Anguelov, D.; Szegedy, C.; Erhan, D.; and Rabinovich, A.
  2015.
\newblock Training Deep Neural Networks on Noisy Labels with Bootstrapping.
\newblock In \emph{International Conference on Learning Representations
  (Workshop)}.

\bibitem[{Shen et~al.(2018)Shen, Lin, Shen, and Reid}]{shen2018bootstrapping}
Shen, T.; Lin, G.; Shen, C.; and Reid, I. 2018.
\newblock Bootstrapping the performance of webly supervised semantic
  segmentation.
\newblock In \emph{Proceedings of the IEEE/CVF Conference on Computer Vision
  and Pattern Recognition}, 1363--1371.

\bibitem[{Shen et~al.(2020)Shen, Ji, Chen, Hong, Zheng, Liu, Xu, and
  Tian}]{shen2020noise}
Shen, Y.; Ji, R.; Chen, Z.; Hong, X.; Zheng, F.; Liu, J.; Xu, M.; and Tian, Q.
  2020.
\newblock Noise-aware fully webly supervised object detection.
\newblock In \emph{Proceedings of the IEEE/CVF Conference on Computer Vision
  and Pattern Recognition}, 11326--11335.

\bibitem[{Simonyan and Zisserman(2014)}]{simonyan2014very}
Simonyan, K.; and Zisserman, A. 2014.
\newblock Very deep convolutional networks for large-scale image recognition.
\newblock \emph{arXiv preprint arXiv:1409.1556}.

\bibitem[{Song et~al.(2022)Song, Kim, Park, Shin, and Lee}]{song2022learning}
Song, H.; Kim, M.; Park, D.; Shin, Y.; and Lee, J.-G. 2022.
\newblock Learning from noisy labels with deep neural networks: A survey.
\newblock \emph{IEEE Transactions on Neural Networks and Learning Systems}.

\bibitem[{Sun et~al.(2017)Sun, Shrivastava, Singh, and
  Gupta}]{sun2017revisiting}
Sun, C.; Shrivastava, A.; Singh, S.; and Gupta, A. 2017.
\newblock Revisiting unreasonable effectiveness of data in deep learning era.
\newblock In \emph{Proceedings of the IEEE/CVF International Conference on
  Computer Vision}, 843--852.

\bibitem[{Sun et~al.(2021)Sun, Yao, Wei, Zhang, Shen, Wu, Zhang, and
  Shen}]{sun2021webly}
Sun, Z.; Yao, Y.; Wei, X.-S.; Zhang, Y.; Shen, F.; Wu, J.; Zhang, J.; and Shen,
  H.~T. 2021.
\newblock Webly supervised fine-grained recognition: Benchmark datasets and an
  approach.
\newblock In \emph{Proceedings of the IEEE/CVF International Conference on
  Computer Vision}, 10602--10611.

\bibitem[{Tanaka et~al.(2018)Tanaka, Ikami, Yamasaki, and
  Aizawa}]{tanaka2018joint}
Tanaka, D.; Ikami, D.; Yamasaki, T.; and Aizawa, K. 2018.
\newblock Joint optimization framework for learning with noisy labels.
\newblock In \emph{Proceedings of the IEEE/CVF Conference on Computer Vision
  and Pattern Recognition}, 5552--5560.

\bibitem[{Tu et~al.(2020)Tu, Niu, Chen, Cheng, and Zhang}]{tu2020learning}
Tu, Y.; Niu, L.; Chen, J.; Cheng, D.; and Zhang, L. 2020.
\newblock Learning from web data with self-organizing memory module.
\newblock In \emph{Proceedings of the IEEE/CVF Conference on Computer Vision
  and Pattern Recognition}, 12846--12855.

\bibitem[{Van~der Maaten and Hinton(2008)}]{van2008visualizing}
Van~der Maaten, L.; and Hinton, G. 2008.
\newblock Visualizing data using t-SNE.
\newblock \emph{Journal of Machine Learning Research}, 9(11).

\bibitem[{Wah et~al.(2011)Wah, Branson, Welinder, Perona, and
  Belongie}]{wah2011caltech}
Wah, C.; Branson, S.; Welinder, P.; Perona, P.; and Belongie, S. 2011.
\newblock The caltech-ucsd birds-200-2011 dataset.

\bibitem[{Wang, Liu, and Tao(2017)}]{wang2017multiclass}
Wang, R.; Liu, T.; and Tao, D. 2017.
\newblock Multiclass learning with partially corrupted labels.
\newblock \emph{IEEE Transactions on Neural Networks and Learning Systems},
  29(6): 2568--2580.

\bibitem[{Wu et~al.(2021)Wu, Wei, Jiang, Mao, Tang, and Li}]{wu2021ngc}
Wu, Z.-F.; Wei, T.; Jiang, J.; Mao, C.; Tang, M.; and Li, Y.-F. 2021.
\newblock NGC: a unified framework for learning with open-world noisy data.
\newblock In \emph{Proceedings of the IEEE/CVF International Conference on
  Computer Vision}, 62--71.

\bibitem[{Xiao et~al.(2015)Xiao, Xia, Yang, Huang, and Wang}]{xiao2015learning}
Xiao, T.; Xia, T.; Yang, Y.; Huang, C.; and Wang, X. 2015.
\newblock Learning from massive noisy labeled data for image classification.
\newblock In \emph{Proceedings of the IEEE/CVF Conference on Computer Vision
  and Pattern Recognition}, 2691--2699.

\bibitem[{Xu et~al.(2016)Xu, Huang, Zhang, and Tao}]{xu2016webly}
Xu, Z.; Huang, S.; Zhang, Y.; and Tao, D. 2016.
\newblock Webly-supervised fine-grained visual categorization via deep domain
  adaptation.
\newblock \emph{IEEE Transactions on Pattern Analysis and Machine
  Intelligence}, 40(5): 1100--1113.

\bibitem[{Yang et~al.(2020)Yang, Feng, Chen, Yan, Zheng, Luo, and
  Zhang}]{yang2020webly}
Yang, J.; Feng, L.; Chen, W.; Yan, X.; Zheng, H.; Luo, P.; and Zhang, W. 2020.
\newblock Webly supervised image classification with self-contained confidence.
\newblock In \emph{European Conference on Computer Vision}, 779--795. Springer.

\bibitem[{Yao et~al.(2020)Yao, Hua, Gao, Sun, Li, and Zhang}]{yao2020bridging}
Yao, Y.; Hua, X.; Gao, G.; Sun, Z.; Li, Z.; and Zhang, J. 2020.
\newblock Bridging the web data and fine-grained visual recognition via
  alleviating label noise and domain mismatch.
\newblock In \emph{Proceedings of the ACM International Conference on
  Multimedia}, 1735--1744.

\bibitem[{Yao et~al.(2017)Yao, Zhang, Shen, Hua, Xu, and
  Tang}]{yao2017exploiting}
Yao, Y.; Zhang, J.; Shen, F.; Hua, X.; Xu, J.; and Tang, Z. 2017.
\newblock Exploiting web images for dataset construction: A domain robust
  approach.
\newblock \emph{IEEE Transactions on Multimedia}, 19(8): 1771--1784.

\bibitem[{Zhang et~al.(2021)Zhang, Bengio, Hardt, Recht, and
  Vinyals}]{zhang2021understanding}
Zhang, C.; Bengio, S.; Hardt, M.; Recht, B.; and Vinyals, O. 2021.
\newblock Understanding deep learning (still) requires rethinking
  generalization.
\newblock \emph{Communications of the ACM}, 64(3): 107--115.

\bibitem[{Zhang et~al.(2020)Zhang, Yao, Liu, Xie, Shu, Zhou, Zhang, Shen, and
  Tang}]{zhang2020web}
Zhang, C.; Yao, Y.; Liu, H.; Xie, G.-S.; Shu, X.; Zhou, T.; Zhang, Z.; Shen,
  F.; and Tang, Z. 2020.
\newblock Web-supervised network with softly update-drop training for
  fine-grained visual classification.
\newblock In \emph{Proceedings of the AAAI Conference on Artificial
  Intelligence}, volume~34, 12781--12788.

\bibitem[{Zhang et~al.(2018)Zhang, Cisse, Dauphin, and
  Lopez-Paz}]{zhang2018mixup}
Zhang, H.; Cisse, M.; Dauphin, Y.~N.; and Lopez-Paz, D. 2018.
\newblock MixUp: Beyond Empirical Risk Minimization.
\newblock In \emph{International Conference on Learning Representations}.

\bibitem[{Zhang and Sabuncu(2018)}]{zhang2018generalized}
Zhang, Z.; and Sabuncu, M. 2018.
\newblock Generalized Cross Entropy Loss for Training Deep Neural Networks with
  Noisy Labels.
\newblock \emph{Advances in Neural Information Processing Systems}, 31.

\bibitem[{Zhou et~al.(2020)Zhou, Pan, Zheng, Xu, and Jin}]{zhou2020large}
Zhou, X.; Pan, P.; Zheng, Y.; Xu, Y.; and Jin, R. 2020.
\newblock Large scale long-tailed product recognition system at alibaba.
\newblock In \emph{Proceedings of ACM International Conference on Information
  \& Knowledge Management}, 3353--3356.

\end{thebibliography}

\end{document}